\documentclass{article} 
\usepackage{iclr2027_conference,times}

\usepackage{amsmath,amsfonts,bm}

\def\eqref#1{equation~\ref{#1}}

\def\1{\bm{1}}

\DeclareMathAlphabet{\mathsfit}{\encodingdefault}{\sfdefault}{m}{sl}
\SetMathAlphabet{\mathsfit}{bold}{\encodingdefault}{\sfdefault}{bx}{n}

\usepackage{hyperref}
\usepackage{url}
\usepackage{amssymb}
\usepackage{enumitem}
\usepackage{booktabs}
\usepackage{multirow} 
\usepackage{graphicx}
\usepackage[table]{xcolor}
\usepackage{array}
\usepackage{float}
\usepackage{wrapfig}
\usepackage[most]{tcolorbox}
\usepackage{listings}
\usepackage{xcolor}

\newtcblisting{promptbox}[1][]{
    enhanced,
    breakable,
    listing only,
    colback=gray!4,
    colframe=gray!45,
    boxrule=0.2pt,
    arc=2pt,
    left=1.5mm,
    right=1.5mm,
    top=0.8mm,
    bottom=0.8mm,
    title=#1,
    fonttitle=\small\bfseries,
    coltitle=black,
    attach boxed title to top left={
        xshift=2mm,
        yshift=-1.5mm
    },
    boxed title style={
        colback=white,
        colframe=gray!45,
        boxrule=0.5pt,
        arc=1pt
    },
    listing options={
        basicstyle=\ttfamily\small,
        columns=fullflexible,
        keepspaces=true,
        breaklines=true,
        showstringspaces=false
    }
}

\title{DataFoundry: Evolving Data Preparators via Recursive Self-Improvement}

\author{
 \textbf{Cehao Yang\textsuperscript{1,2,3}\footnotemark[1]},
 \textbf{ Xiaojun Wu\textsuperscript{1,2,3}\footnotemark[1] },
 \textbf{ Xueyuan Lin\textsuperscript{1,2}\footnotemark[1]},\\
 \textbf{ Chengjin Xu\textsuperscript{1,3}}\footnotemark[2],
 \textbf{ Xuhui Jiang\textsuperscript{1,3}},
 \textbf{ Hui Xiong\textsuperscript{2}}\footnotemark[2],
 \textbf{ Jian Guo\textsuperscript{1}}\footnotemark[2]
\\
 \textsuperscript{1}IDEA Research, International Digital Economy Academy
 \\
 \textsuperscript{2}Hong Kong University of Science and Technology (Guangzhou)
\\
 \textsuperscript{3}DataArc Tech Ltd.
 \\
 \texttt{\{cyang289,xwu647,xlin058\}@connect.hkust-gz.edu.cn},
 \\
 \texttt{xionghui@ust.hk}, \texttt{\{xuchengjin,jiangxuhui,guojian\}@idea.edu.cn}
}

\iclrfinalcopy 
\begin{document}

\maketitle
\footnotetext[1]{* means equal contribution.}
\footnotetext[2]{† means corresponding authors.}

\begin{abstract}
Domain adaptation of large language models increasingly depends on constructing high-quality training data, yet existing data-preparation pipelines typically address quality only after generation through post-hoc filtering. This creates a fundamental mismatch: data-quality issues often originate from the construction process itself, while quality control is applied only to its outputs. We introduce \textsc{DataFoundry}, a framework for \textbf{evolving data preparators through recursive self-improvement} before large-scale data production. \textsc{DataFoundry} represents a data preparator as an evolvable runtime specification and instantiates its evolution with a \textsc{Skills-as-Modules} architecture, in which a central \textsc{Controller} orchestrates modular skills to compile executable runtimes, diagnose deficiencies on small pilot sets using domain-appropriate criteria, and translate diagnostic feedback into adapters that revise individual preparation components while preserving stable interfaces. We evaluate \textsc{DataFoundry} on DataPrep-Bench across mathematics, finance, law, and medicine, and find that recursively evolved preparators produce training data with higher downstream utility than baselines. Experiments across different backbones further demonstrate that these improvements are not tied to a particular model, while analyses and case studies further reveal the framework's optimization dynamics and illustrate how its evolution unfolds in practice.
\end{abstract}

\section{Introduction}
\label{sec:introduction}

Large language models adapted to specialized domains rely on the effective preparation of domain knowledge as training data~\citep{ling2025domain, liang2026dataprepbenchbenchmarkingllmstraining}. Instead of directly training on raw corpora, the adaptation typically involves restructuring them into examples that better align with the target tasks and desired model behaviors~\citep{lu2025fine, nayak2024learning}. Synthetic instruction generation has made this process easier to scale. Early pipelines bootstrap instructions and responses from model generations~\citep{wang2023selfinstructaligninglanguagemodels,ding2023enhancingchatlanguagemodels, xu2025wizardlmempoweringlargepretrained}, then remove invalid, redundant, or low-quality examples~\citep{li2025infinityinstructscalinginstruction, yang2026select2reason}. 

\begin{wrapfigure}{r}{0.5\textwidth}
    \centering
    \vspace{-5pt}
    \includegraphics[width=0.5\textwidth]{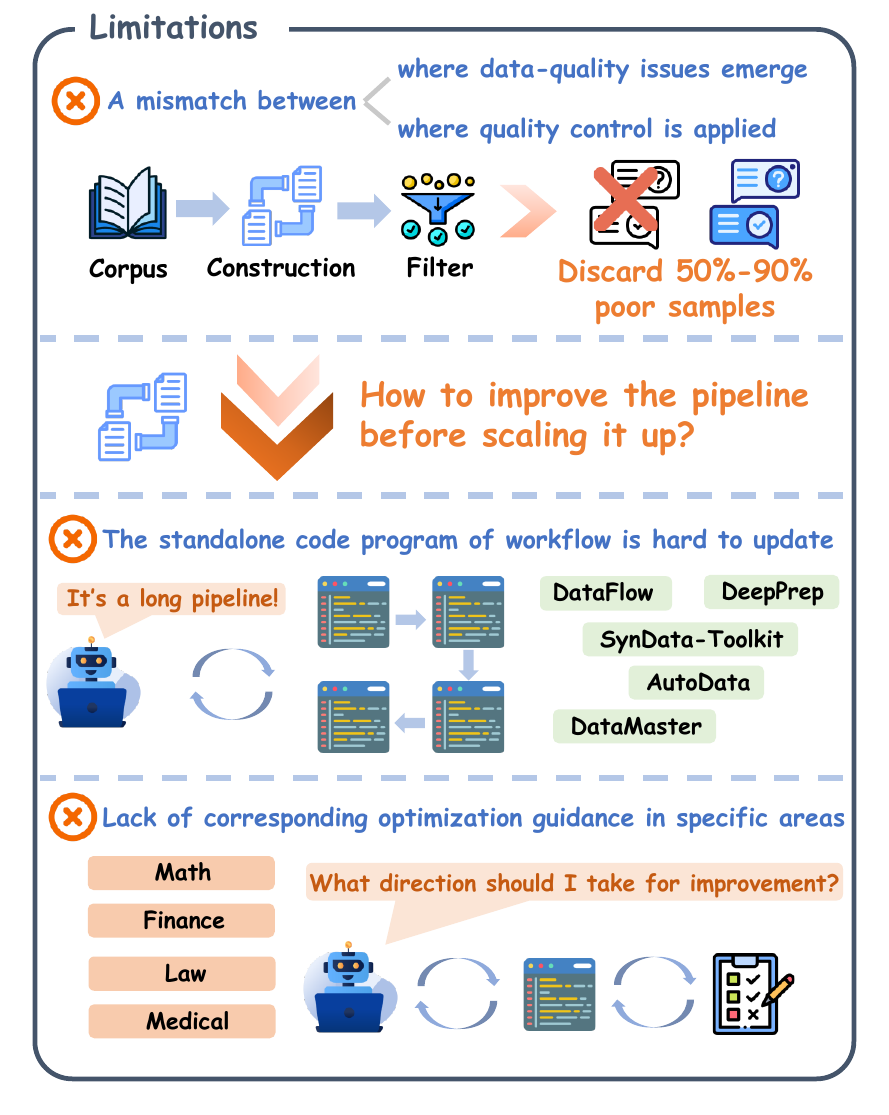}
    \caption{A mismatch between where data-quality issues emerge and where quality control is applied during data preparation. Challenges exist when optimizing the pipeline before scaling it up.}
    \vspace{-5pt}
    \label{fig:limitation}
\end{wrapfigure}

However, existing pipelines exhibit a fundamental mismatch between where quality issues emerge and where quality control is applied: \textit{quality issues can arise during the construction process, whereas quality control is typically applied only to the resulting samples}. As shown in Figure~\ref{fig:limitation}, these pipelines often rely on post-hoc filtering to discard large fractions of generated data~\citep{xu2025magpie, yu2025cot, li2024self}, rather than correcting the upstream processes that repeatedly produce poor examples. This mismatch suggests that improving data quality requires \textbf{optimizing the preparation pipeline itself before scaling it up}.

Addressing this mismatch requires shifting quality intervention upstream, but doing so is far from straightforward. First, data construction often involves long, multi-stage workflows implemented as standalone code~\citep{fan2026deepprepllmpoweredagenticautonomous, shi2026dataarc, kulikov2026autodataagenticdatascientist, du2026datamasterdatacentricautonomousai}, making it difficult to \textbf{diagnose and incorporate iterative improvements}. Second, even when problematic stages can be identified, there is \textbf{limited actionable guidance on what should be optimized and how}, particularly across domains with different requirements, such as mathematics, finance, law, and medicine~\citep{ling2025domain}. These challenges motivate a systematic approach that can trace quality issues back to their construction processes and provide targeted, domain-aware guidance for improving the pipeline before the final large-scale data preparation.

To this end, we introduce \textsc{DataFoundry}, a framework for \textbf{evolving data preparators through recursive self-improvement} before large-scale data production. It represents it as an evolvable runtime specification. We instantiate this process with a \textsc{Skills-as-Modules} architecture, in which a central \textsc{Controller} orchestrates modular skills for compiling the current specification into an executable runtime, diagnosing deficiencies on small pilot sets using domain-appropriate criteria, and translating the resulting feedback into adapters that revise the preparator while preserving stable component interfaces. Through this recursive process, \textsc{DataFoundry} progressively specializes a general data preparator to the requirements of each target domain, and finally compiles the evolved specification into a domain-adapted runtime for full-scale supervised fine-tuning data production.

We evaluate \textsc{DataFoundry} across four specialized domains, including \textit{Math}, \textit{Finance}, \textit{Law}, and \textit{Medical}, and show that its recursively evolved data preparators consistently improve the downstream utility of the prepared training data. Beyond the main results with Qwen2.5-7B, we conduct ablation studies with different downstream base models to examine whether the gains generalize beyond a particular model. We further vary the self-improvement rounds to study how performance evolves and to characterize the growth potential of the proposed framework. Finally, through case studies, we trace the changes introduced during recursive self-improvement to individual pipeline components, providing a component-level explanation of how \textsc{DataFoundry} progressively adapts the data-production process to domain-specific requirements.

\section{Related Work}
\label{sec:related}

\subsection{LLM Data Construction}
Early data construction pipelines bootstrap instructions from model generations using filters to control noise~\citep{wang2023selfinstructaligninglanguagemodels,ding2023enhancingchatlanguagemodels}. Subsequent work increases task complexity through iterative instruction rewriting~\citep{xu2025wizardlmempoweringlargepretrained}, or couples large-scale synthesis to improve instruction-following quality~\citep{li2025infinityinstructscalinginstruction}. Recent systems elevate data preparation from a fixed generator to a modular workflow~\citep{shi2026dataarc, fan2026deepprepllmpoweredagenticautonomous, liang2025dataflowllmdrivenframeworkunified}. DataMaster searches over cumulative data states with downstream training feedback, and Autodata iteratively creates, inspects, evaluates, and revises synthetic-data recipes~\citep{du2026datamasterdatacentricautonomousai,kulikov2026autodataagenticdatascientist}. Some works attempt to optimize the prompt iteratively for data generation~\citep{zeng2024automatic}, in contrast, \textsc{DataFoundry} recursively adapts the executable data construction runtime itself using contextualized assessment, then performs full-scale construction on evolved runtime.

\subsection{Agent Skills and Evolution}
Agent skills externalize procedural knowledge as composable packages of instructions, code, and resources, providing a portable adaptation surface~\citep{xu2026agentskillslargelanguage}. One line of work turns experience into reusable skills by distilling trajectories or extracting from recurring interactrions~\citep{xia2026skillrlevolvingagentsrecursive,yang2026autoskillexperiencedrivenlifelonglearning,ma2026skillclawletskillsevolve}. A complementary line directly optimizes skill artifacts. CoEvoSkills couples multi-file skill generation with a co-evolving surrogate verifier~\citep{zhang2026coevoskillsselfevolvingagentskills}; Skill-R1 trains a lightweight skill generator with intra- and inter-generation rewards while freezing the task model~\citep{vishe2026skillr1agentskillevolution}; and SkillOpt casts skill revision as bounded text-space optimization with held-out tests~\citep{yang2026skilloptexecutivestrategyselfevolving}. Unlike existing methods, \textsc{DataFoundry} evolves the skill with domain-specific changes isolated as inspectable specification adapters.

\subsection{Recursive Self-Improvement}
Recursive self-improvement (RSI) spans qualitatively different update surfaces, from bounded output refinement and self-training to evaluator improvement, harness modification, and autonomous research loops~\citep{lee2026meta, chen2026recursiveselfimprovementaibounded, lee2026recursiveharnessselfimprovement}. Experience-centric accounts accordingly view the runtime harness as the infrastructure that captures trajectories and feedback and routes them into mutable skills, memory, environments, or parameters~\citep{jiang2026selfimprovingagents}. Harness engineering makes this external update surface concrete through explicit workflows, persistent state, context management, evaluation, and permission boundaries~\citep{weng2026harness}. \textsc{DataFoundry} instantiates a bounded RSI loop specifically for data preparation: multi-criterion assessment grounds each adapter update, while the baseline runtime specification constrains potential drift.

\section{DataFoundry}
\label{sec:method}

\subsection{Background: Data Preparation}
\label{sec:background}

\paragraph{Problem Formulation.}
Let $d$ denote a target domain and let $\mathcal{K}_d$ denote its knowledge corpus, where each $k \in \mathcal{K}_d$ is a domain-relevant knowledge source, such as a textbook, novel, or other document. A data-construction runtime $\mathcal{R}$ transforms the domain corpus into a usable domain-specific supervised fine-tuning dataset:
\begin{equation}
\mathcal{D}^{\mathrm{SFT}}_d
=
\mathcal{R}(\mathcal{K}_d)
=
\{z_i\}_{i=1}^{N_d},
\quad
z_i=(\iota_i,\rho_i),
\label{eq:data_preparation}
\end{equation}
where $N_d$ denotes the realized data volume yielded, and $\iota_i$ and $\rho_i$ denote the instruction and response of a training sample. Given a base model $f_{\theta_0}$, the prescribed fine-tuning procedure $\mathcal{A}$ uses the prepared fine-tuning dataset $\mathcal{D}^{\mathrm{SFT}}_d$ to train a downstream domain model $f_{\theta_d}$:
\begin{equation}
f_{\theta_d}
=
\mathcal{A}(f_{\theta_0},\mathcal{D}^{\mathrm{SFT}}_d).
\label{eq:downstream_training}
\end{equation}

\paragraph{Construction-then-Filtering.}
A common instantiation follows a construction-then-filtering paradigm, in which candidate training data are first constructed and subsequently filtered. The data-construction runtime $\mathcal{R}$ transforms the corpus $\mathcal{K}_d$ into a candidate training data pool:
\begin{equation}
\widetilde{\mathcal{D}}^{\mathrm{SFT}}_d
=
\mathcal{R}(\mathcal{K}_d)
=
\{z_i\}_{i=1}^{\widetilde{N}_d},
\label{eq:candidate_data_construction}
\end{equation}
where $\widetilde{N}_d$ denotes the volume of the candidate pool. A data filter $\mathcal{F}$ evaluates the quality of candidate samples and selects a subset with high value for final downstream fine-tuning, written as
\begin{equation}
\mathcal{D}^{\mathrm{SFT}}_d
=
\mathcal{F}(
\widetilde{\mathcal{D}}^{\mathrm{SFT}}_d)
=
\{z_i\}_{i=1}^{N_d},
\quad
N_d \leq \widetilde{N}_d.
\label{eq:data_filtering}
\end{equation}

\begin{figure}
\centering
\includegraphics[width=\textwidth]{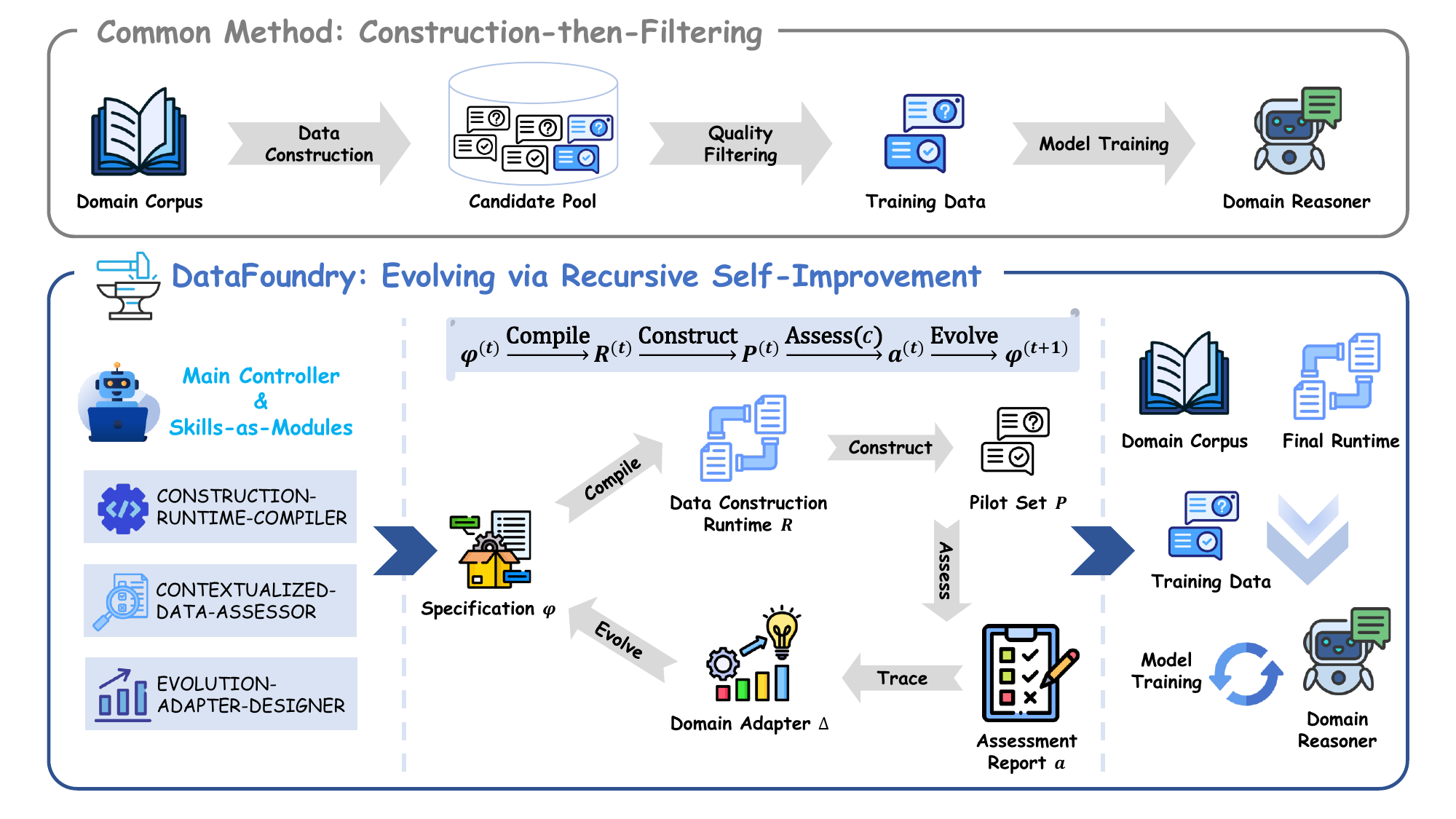}
\caption{The overall framework of \textsc{DataFoundry}. Unlike the common method that constructs first then filters, our \textsc{DataFoundry} evolves the data preparator via recursive self-improvement, prior to final data production.}
\label{fig:main_pipeline}
\end{figure}

\subsection{Inspiration: Evolving via Recursive Self-Improvement}
\label{sec:evolving_rsi}

To address the challenges mentioned above, we shift adaptation upstream from filtering the outputs to improving the preparator itself, using \textit{recursive self-improvement}~\citep{chen2026recursiveselfimprovementaibounded} before final production. We begin with an initial runtime specification $\phi_\text{base}$ that describes the general workflow of data production. A runtime compiler translates $\phi_\text{base}$ into an online construction runtime:
\begin{equation}
\widetilde{\mathcal{R}}
=
\operatorname{Compile}(\phi_\text{base}).
\label{eq:initial_runtime_compilation}
\end{equation}
For a target domain $d$, a contextualized quality assessor selects domain-specific criteria $\mathcal{C}_d$ to examine data quality from multiple perspectives. At iteration $t$, the current specification $\phi_d^{(t)}$ is compiled into a runtime $\widetilde{\mathcal{R}}_d^{(t)}$, which then constructs a pilot set with a small generation budget $\widehat{N}$:
\begin{equation}
\mathcal{P}_d^{(t)}=\widetilde{\mathcal{R}}_d^{(t)}(\mathcal{K}_d;\widehat{N}).
\label{eq:pilot_set}
\end{equation}
The assessor evaluates this pilot set and returns the report:
\begin{equation}
a_d^{(t)}=\operatorname{Assess}(\mathcal{P}_d^{(t)};\mathcal{C}_d).
\end{equation}
An evolution operator then updates the specification using this feedback:
\begin{equation}
\phi_d^{(t+1)}
=
\operatorname{Evolve}(\phi_d^{(t)},a_d^{(t)}),
\quad
\phi_d^{(0)}=\phi_\text{base}.
\label{eq:recursive_runtime_evolution}
\end{equation}
In summary, a \textit{recursive self-improvement} process is formalized in the following loop:
\begin{equation}
    \phi_d^{(t)}
    \xrightarrow{\;\operatorname{Compile}\;}
    \widetilde{\mathcal{R}}_d^{(t)}
    \xrightarrow{\;\mathcal{K}_d,\widehat{N}\;}
    \mathcal{P}_d^{(t)}
    \xrightarrow{\;\operatorname{Assess}(\cdot;\mathcal{C}_d)\;}
    a_d^{(t)}
    \xrightarrow{\;\operatorname{Evolve}\;}
    \phi_d^{(t+1)}.
    \label{eq:evolution_loop}
\end{equation}
After $T$ rounds, the evolved specification is compiled into the final domain-adapted runtime for full-scale production of the fine-tuning dataset:
\begin{equation}
\mathcal{R}_d^{(T)}
=
\operatorname{Compile}(\phi_d^{(T)}),
\quad
\mathcal{D}^{\mathrm{SFT}}_d
=
\mathcal{R}_d^{(T)}(\mathcal{K}_d)
=
\{z_i\}_{i=1}^{N_d}.
\label{eq:evolved_full_data_production}
\end{equation}

\subsection{Framework: Controller and Skills-as-Modules}
\label{sec:framework_overview}

\textsc{DataFoundry} implements the RSI process above through a \textsc{Skills-as-Modules} architecture, in which a main \textsc{Controller} driven by a coding agent $\mathcal{G}$ orchestrates three modular skills to compile, assess, and evolve the data-construction runtime specification $\phi$:
\begin{itemize}[leftmargin=2em]
    \item \textsc{\textbf{Construction-Runtime-Compiler}} realizes $\operatorname{Compile}$, in which $\mathcal{G}$ translates the specification $\phi$ into an executable data-construction runtime $\mathcal{R}$.
    \item \textsc{\textbf{Contextualized-Data-Assessor}} realizes $\operatorname{Assess}$, in which $\mathcal{G}$ selects a set of domain-appropriate criteria $\mathcal{C}_d$ and applies them to the pilot set $\mathcal{P}_d$.
    \item \textsc{\textbf{Evolution-Adapter-Designer}} realizes $\operatorname{Evolve}$, in which $\mathcal{G}$ attributes diagnosed deficiencies to components and expresses the required changes as an adapter for $\phi_d^{(t)}$.
\end{itemize}

The following sections describe the concrete realization of compilation, assessment, and evolution.

\subsubsection{Data-Construction Runtime Compilation}
\label{sec:runtime_compilation}

This skill defines the initial runtime specification $\phi$ including four stable component interfaces:
\begin{itemize}[leftmargin=2em]
    \item \textbf{\texttt{corpus\_chunk\_manager}} samples source chunks from $\mathcal{K}_d$ to provide the knowledge context for the construction of sample $z=(\iota,\rho)$;
    \item \textbf{\texttt{instruction\_builder}} converts the sampled context into a domain-relevant instruction $\iota$ in the required format;
    \item \textbf{\texttt{response\_generator}} produces a formatted response $\rho$ conditioned on the instruction and its source context; and
    \item \textbf{\texttt{batch\_generation\_service}} concurrently invokes external models to produce samples.
\end{itemize}
The compiler additionally provides engineering constraints like smoke tests and structured logging, which make each compiled runtime stable and observable.

\subsubsection{Contextualized Pilot-Set Assessment}
\label{sec:criteria_assessment}

This skill defines the data quality assessment equipped with a library $\mathcal{L}_{\mathrm{crit}}$ of candidate assessment criteria. The candidate criteria cover several complementary quality signals:

\begin{itemize}[leftmargin=2em]
\item \textbf{\texttt{instruction\_alignment}}: whether the response directly, accurately, and sufficiently addresses the given instruction and its intended task;
\item \textbf{\texttt{self\_containment}}: whether the sample is complete and understandable without relying on missing context or unspecified external information;
\item \textbf{\texttt{information\_density}}: whether the response contains sufficient domain-relevant and useful information while avoiding excessive redundancy or unnecessary verbosity;
\item \textbf{\texttt{factual\_grounding}}: whether the response is well supported by the provided source context and avoids unsupported, inconsistent, or fabricated claims;
\item \textbf{\texttt{domain\_utility}}: whether the sample captures informative domain knowledge, concepts, and reasoning patterns that are useful for downstream domain adaptation; and
\item \textbf{\texttt{numerical\_correctness}}: whether numerical values, calculations, quantitative relationships, and arithmetic reasoning in the response are accurate and internally consistent.
\end{itemize}

Given a domain $d$, the domain-specific criteria $\mathcal C_d\subseteq\mathcal L_{\mathrm{crit}}$ are selected contextually and applied to assess the pilot set $\mathcal{P}_d^{(t)}$, and the findings given by $\mathcal{G}$ are aggregated into the assessment report $a_d^{(t)}$.

\subsubsection{Adapter-Based Specification Evolution}
\label{sec:adapter_evolution}

This skill takes the specification $\phi_\text{base}$, the adapter $\Delta_d^{(t-1)}$ and the report $a_d^{(t)}$, then traces the diagnosed deficiencies to components. It then expresses the recommended modifications as:
\begin{equation}
\Delta_d^{(t)}
=
\operatorname{Trace}(\phi_\text{base}, \Delta_d^{(t-1)}, a_d^{(t)}).
\label{eq:adapter_generation}
\end{equation}

The adapter may revise component-level policies, prompts, configurations, or coordination logic while preserving the stable interfaces enforced by the compiler. In particular, it can modify the behavior of components such as \texttt{corpus\_chunk\_manager}, \texttt{instruction\_builder}, and \texttt{response\_generator}, thereby addressing deficiencies in the construction process.

The \textsc{Controller} applies the adapter to obtain the specification used in the next pilot iteration:
\begin{equation}
\phi_d^{(t+1)}
=
\operatorname{Adapt}(\phi_\text{base}, \Delta_d^{(t)}).
\label{eq:adapter_adaption}
\end{equation}

By isolating domain-specific changes in $\Delta_d^{(t)}$ while preserving the baseline $\phi_\text{base}$, it mitigates iterative updates from accumulating uncontrolled modifications that may lead to specification collapse.
\section{Experiments}
\label{sec:experiments}

\subsection{Implementation Details}

We conduct experiments on a Linux server equipped with (8$\times$) NVIDIA H800 GPUs. We use GPT-5.6 Sol~\footnote{https://developers.openai.com/api/docs/models} as the backbone model for the coding agent and Codex CLI~\footnote{https://learn.chatgpt.com/docs/codex/cli} as its execution interface. We follow the domain corpus used in~\citep{liang2026dataprepbenchbenchmarkingllmstraining} with 136 math documents, 80 finance documents, 53 law documents and 154 medical documents. The details of our skills-as-modules framework are provided in Appendix~\ref{app:skills}, and the downstream SFT setup are provided in Appendix~\ref{app:training}.

\subsection{Evaluation Settings}

The evaluation benchmarks and existing experimental results are adopted from~\citep{liang2026dataprepbenchbenchmarkingllmstraining}, covering four domains and various sub-tasks: (1) \textbf{Math}: GSM8K, AMC 2023, AIME 2024, MinervaMath, Math-5000, Gaokao 2024, OlympiadBench; (2) \textbf{Finance}: CPA-KQA, FinEval-KR, XFinBench; (3) \textbf{Law}: LegalBench, LexGLUE; (4) \textbf{Medical}: MedCaseReasoning, MedMCQA, MedR-Bench. The details of the four benchmark datasets are provided in Appendix~\ref{app:evaluation}.

All baselines and our \textsc{DataFoundry} are fine-tuned on Dolly-15k~\footnote{https://huggingface.co/datasets/databricks/databricks-dolly-15k} (a general instruction-following corpus) jointly with the domain-specific dataset produced by the named generator. We fix the number of RSI rounds at two, as we observe that the pilot set assessment results typically begin to deteriorate after two rounds. The implementation of baselines are provided in Appendix~\ref{app:baselines}.

\begin{table}[t]
\caption{Performance of Qwen2.5-7B on Math and Finance benchmarks after fine-tuning on Dolly-15k jointly with the datasets synthesized by different generators.}
\label{tab:synthetic_data_mf_qwen}
\centering
\scriptsize
\setlength{\tabcolsep}{2pt}

\begin{tabular*}{\columnwidth}{
    @{\extracolsep{\fill}}
    l|cccccccc|cccc
    @{}
}
\toprule
\multirow{2}{*}{\small\textbf{Training Data Generator}} 
& \multicolumn{8}{c|}{\small\textbf{Math}} 
& \multicolumn{4}{c}{\small\textbf{Finance}} \\
\cmidrule(lr){2-9}\cmidrule(lr){10-13}
& GSM8K & AMC23 & AIME24 & MM & OB & M5000 & GK24 & Avg 
& CKQA & FEKR & XFB & Avg \\
\midrule
Dolly-15k only & 69.9 & 17.5 & 0.0 & 10.7 & 10.7 & 39.8 & 16.5 & 23.6 & 57.6 & 59.4 & 56.3 & 57.8 \\
\midrule
\rowcolor[rgb]{.867, .922, .969}
\multicolumn{13}{c}{\textit{\small\textbf{DataFlow-based Generators}}} \\
\midrule
DataFlow & 56.5 & 7.5 & 0.0 & 7.7 & 7.9 & 27.4 & 17.6 & 17.8 & 51.0 & 54.5 & 59.3 & 54.9 \\
DataFlow-Skill & 56.7 & 10.0 & 0.0 & 8.8 & 6.2 & 22.8 & \textbf{28.6} & 19.0 & 60.0 & 65.4 & \textbf{68.9} & 64.8 \\
\midrule
\rowcolor[rgb]{.867, .922, .969}
\multicolumn{13}{c}{\textit{\small\textbf{LLM-based Generators}}} \\
\midrule
Claude Opus 4.6 & 68.8 & 12.5 & 0.0 & 8.8 & 9.8 & 35.5 & 14.3 & 21.4 & 37.6 & 39.6 & 55.9 & 44.4 \\
Gemini 3.0 Pro & 72.1 & 20.0 & 0.0 & 10.7 & 11.4 & 37.9 & 15.4 & 23.9 & 48.6 & 49.5 & 58.4 & 52.2 \\
GPT-5.2 & 66.7 & 17.5 & 0.0 & 7.7 & 11.1 & 32.7 & 14.3 & 21.4 & 47.1 & 43.6 & 53.6 & 48.1 \\
\midrule
\rowcolor[rgb]{.867, .922, .969}
\multicolumn{13}{c}{\textit{\small\textbf{Agent-based Generators}}} \\
\midrule
Qwen3.5-Plus & 72.7 & 22.5 & 3.3 & 11.0 & 11.6 & 38.7 & 16.5 & 25.2 & 48.1 & 49.5 & 55.6 & 51.1 \\
GLM-4.7 & 71.4 & 22.5 & 3.3 & 9.6 & 11.6 & 37.9 & 20.9 & 25.3 & 51.4 & 55.4 & 53.1 & 53.3 \\
Claude Opus 4.6 & 69.4 & 10.0 & 0.0 & 11.0 & 8.9 & 33.8 & 19.8 & 21.8 & 32.4 & 39.6 & 53.8 & 41.9 \\
Gemini 3.0 Pro & 70.1 & 15.0 & 0.0 & 8.8 & 10.8 & 35.7 & 20.9 & 23.0 & 53.3 & 52.5 & 55.4 & 53.7 \\
GPT-5.2 & 69.6 & 25.0 & 3.3 & 8.8 & 9.9 & 35.6 & 26.4 & 25.5 & 39.1 & 39.6 & 51.0 & 43.2 \\
GPT-5.3-codex & 70.8 & 15.0 & 0.0 & 11.0 & 11.7 & 38.5 & 13.2 & 22.9 & 58.6 & 63.4 & 55.9 & 59.3 \\
GPT-5.6 Sol & 71.8 & 22.5 & 3.3 & 11.4 & 11.7 & 37.5 & 19.8 & 25.4 & 50.5 & 51.5 & 53.1 & 51.7 \\
\midrule
\rowcolor[rgb]{.867, .922, .969}
\multicolumn{13}{c}{\textit{\small\textbf{Skill-based Generators}}} \\
\midrule
DataPrep-Skill~(Opus 4.6) & 72.6 & 17.5 & 3.3 & \textbf{14.0} & 11.1 & 36.8 & 13.2 & 24.1 & 57.6 & 53.5 & 55.4 & 55.5 \\
DataPrep-Skill~(GPT-5.6 Sol) & 71.3 & 17.5 & 3.3 & 9.6 & 11.6 & 40.7 & 20.9 & 25.0 & 49.0 & 64.4 & 66.2 & 59.9 \\
\textbf{DataFoundry~(GPT-5.6 Sol)} & \textbf{75.1} & \textbf{27.5} & \textbf{6.7} & 11.4 & \textbf{13.0} & \textbf{44.7} & 23.1 & \textbf{28.8} & \textbf{61.4} & \textbf{71.3} & 65.5 & \textbf{66.1} \\
\bottomrule
\end{tabular*}
\end{table}

\subsection{Main Results}

\paragraph{\textsc{DataFoundry} achieves the best overall performance across all domains.} As shown in Tables~\ref{tab:synthetic_data_mf_qwen} and~\ref{tab:synthetic_data_lm_qwen}, \textsc{DataFoundry} attains average scores of 28.8, 66.1, 78.6, and 46.0 on Math, Finance, Law, and Medical, respectively, outperforming the strongest competing generator by 3.3, 1.3, 1.4, and 2.2 points. The results demonstrate the effectiveness and generality of recursive self-improvement on data-preparation process. While \textsc{DataFoundry} does not necessarily achieve the highest score on every individual sub-task, it delivers the strongest domain-level averages, indicating that its advantage comes from adapting the preparation runtime to domain-specific needs and producing utility training data.

\paragraph{\textsc{DataFoundry} derives its gains from the framework rather than the backbone model alone.}
To isolate the contribution of our framework from that of the backbone model, we compare \textsc{DataFoundry} with both the \textit{Agent-based Generator} and the \textit{Skill-based Generator} instantiated with the same GPT-5.6 Sol backbone. Compared with the GPT-5.6 Sol Agent-based Generator, \textsc{DataFoundry} improves the domain-level average by 3.4 points on Math, 14.4 points on Finance, 4.8 points on Law, and 11.1 points on Medical. Moreover, compared with the GPT-5.6 Sol Skill-based Generator (DataPrep-Skill), \textsc{DataFoundry} achieves further gains of 3.8, 6.2, 7.9, and 5.9 points across the four domains, respectively. These controlled comparisons demonstrate that the improvements cannot be attributed merely to a stronger backbone, agentic execution, or access to reusable skills, rather, they highlight the effectiveness of \textsc{DataFoundry}'s framework-level design.

\begin{table}[t]
\caption{Performance of Qwen2.5-7B on Law and Medical benchmarks after fine-tuning on Dolly-15k jointly with the datasets synthesized by different generators.}
\label{tab:synthetic_data_lm_qwen}
\centering
\scriptsize
\setlength{\tabcolsep}{6pt}

\begin{tabular*}{\columnwidth}{
    @{\extracolsep{\fill}}
    l|ccc|cccc
    @{}
}
\toprule
\multirow{2}{*}{\small\textbf{Training Data Generator}} 
& \multicolumn{3}{c|}{\small\textbf{Law}} 
& \multicolumn{4}{c}{\small\textbf{Medical}} \\
\cmidrule(lr){2-4}\cmidrule(lr){5-8}
& LegalBench & LexGLUE & Avg 
& MedCaseReasoning & MedMCQA & MedR-Bench & Avg \\
\midrule
Dolly-15k only & 86.9 & 62.0 & 74.5 & 13.6 & 27.4 & 67.8 & 36.3 \\
\midrule
\rowcolor[rgb]{.867, .922, .969}
\multicolumn{8}{c}{\textit{\small\textbf{DataFlow-based Generators}}} \\
\midrule
DataFlow & 89.7 & 64.8 & 77.2 & 9.9 & 29.4 & 63.6 & 34.3 \\
DataFlow-Skill & \textbf{92.0} & 57.4 & 74.7 & 11.9 & 24.1 & 65.6 & 33.9 \\
\midrule
\rowcolor[rgb]{.867, .922, .969}
\multicolumn{8}{c}{\textit{\small\textbf{LLM-based Generators}}} \\
\midrule
Claude Opus 4.6 & 88.0 & 63.2 & 75.6 & 13.9 & 8.1 & 66.8 & 29.6 \\
Gemini 3.0 Pro & 85.9 & 63.5 & 74.7 & 12.2 & 6.0 & 69.1 & 29.1 \\
GPT-5.2 & 89.2 & 60.9 & 75.0 & 10.6 & 23.3 & 66.0 & 33.3 \\
\midrule
\rowcolor[rgb]{.867, .922, .969}
\multicolumn{8}{c}{\textit{\small\textbf{Agent-based Generators}}} \\
\midrule
Qwen3.5-Plus & 90.2 & 61.0 & 75.6 & 16.5 & 16.5 & 68.3 & 33.8 \\
GLM-4.7 & 84.8 & 61.4 & 73.1 & 15.2 & 10.6 & 66.8 & 30.9 \\
Claude Opus 4.6 & 65.2 & 48.8 & 57.0 & 16.6 & 50.7 & 54.3 & 40.5 \\
Gemini 3.0 Pro & 57.5 & 55.5 & 56.5 & 15.8 & \textbf{52.3} & 63.4 & 43.8 \\
GPT-5.2 & 61.7 & 51.9 & 56.8 & 14.7 & 45.0 & 59.0 & 39.6 \\
GPT-5.3-codex & 87.8 & 63.4 & 75.6 & 13.6 & 20.3 & 70.0 & 34.6 \\
GPT-5.6 Sol & 85.0 & 62.6 & 73.8 & 11.3 & 27.8 & 65.6 & 34.9 \\
\midrule
\rowcolor[rgb]{.867, .922, .969}
\multicolumn{8}{c}{\textit{\small\textbf{Skill-based Generators}}} \\
\midrule
DataPrep-Skill~(Opus 4.6) & 85.5 & 61.7 & 73.6 & 9.9 & 15.3 & 65.4 & 30.2 \\
DataPrep-Skill~(GPT-5.6 Sol) & 82.6 & 58.7 & 70.7 & 15.8 & 39.8 & 64.6 & 40.1 \\
\textbf{DataFoundry~(GPT-5.6 Sol)} & 87.4 & \textbf{69.8} & \textbf{78.6} & \textbf{20.2} & 47.0 & \textbf{70.8} & \textbf{46.0} \\
\bottomrule
\end{tabular*}
\end{table}

\begin{figure}
\centering
\includegraphics[width=\textwidth]{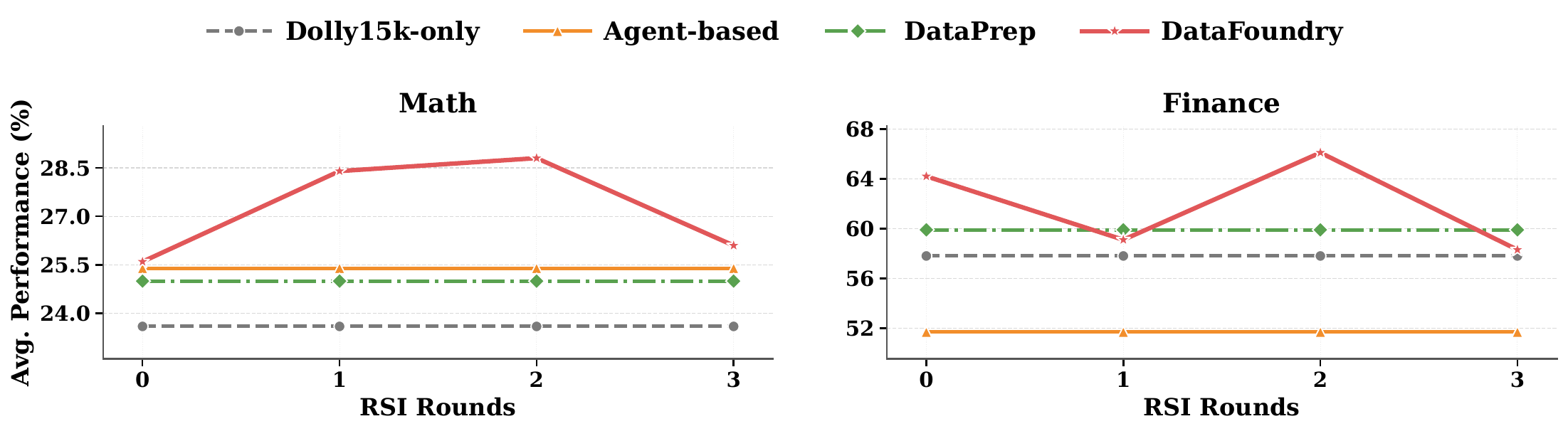}
\caption{Performance on Math and Finance across different RSI rounds using GPT-5.6 Sol as the backbone. \textsc{DataFoundry} generally improves through recursive evolution and achieves its best overall performance after two rounds, while further evolution leads to performance degradation.}
\label{fig:RSI_round}
\end{figure}
\section{Analysis}
\label{sec:analysis}

\subsection{Ablation Studies}

\paragraph{\textsc{DataFoundry} benefits from recursive self-improvement, but excessive evolution can cause degradation.}
As shown in Figure~\ref{fig:RSI_round}, increasing the number of RSI rounds generally improves the downstream utility of the data synthesized by \textsc{DataFoundry}, with the strongest overall performance reached after two rounds of evolution. On Math, the average score increases from 25.6 at Round 0 to 28.4 and 28.8 at Rounds 1 and 2, respectively, before dropping to 26.1 at Round 3. Finance exhibits a less monotonic trajectory, but similarly reaches its best performance at Round 2 (66.1) and declines substantially to 58.3 after an additional round. These results suggest that recursive optimization is beneficial up to a certain point, while excessive self-improvement may introduce undesirable changes to the data-preparation runtime and lead to partial system collapse.

\begin{wraptable}{r}{0.55\columnwidth}
\vspace{-25pt}
\caption{Performance of Llama-3.1-8B.}
\label{tab:synthetic_data_llama}
\centering
\scriptsize
\setlength{\tabcolsep}{2pt}

\begin{tabular*}{\linewidth}{
    @{\extracolsep{\fill}}
    l|cccc
    @{}
}
\toprule
\small\textbf{Data Generator}
& \small\textbf{Math}
& \small\textbf{Finance}
& \small\textbf{Law}
& \small\textbf{Medical} \\
\midrule

Dolly-15k only
& 11.7 & 15.1 & 69.5 & 20.4 \\

\midrule
\rowcolor[rgb]{.867, .922, .969}
\multicolumn{5}{c}{\textit{\small\textbf{DataFlow-based Generators}}} \\
\midrule

DataFlow
& 5.8 & 31.4 & 72.7 & 29.9 \\

DataFlow-Skill
& 6.1 & 36.5 & \textbf{73.1} & 36.0 \\

\midrule
\rowcolor[rgb]{.867, .922, .969}
\multicolumn{5}{c}{\textit{\small\textbf{LLM-based Generators}}} \\
\midrule

Claude Opus 4.6
& 9.3 & 28.4 & 71.0 & 36.1 \\

Gemini 3.0 Pro
& 11.0 & 31.2 & 74.0 & 33.4 \\

GPT-5.2
& 9.2 & 27.4 & 76.4 & 32.4 \\

\midrule
\rowcolor[rgb]{.867, .922, .969}
\multicolumn{5}{c}{\textit{\small\textbf{Agent-based Generators}}} \\
\midrule
Claude Opus 4.6
& 6.7 & 28.9 & 69.8 & 32.7 \\

Gemini 3.0 Pro
& 10.9 & 29.3 & 73.5 & 34.2 \\

GPT-5.2
& 10.0 & 29.5 & 72.9 & 32.7 \\

GPT-5.6 Sol
& 9.7 & 30.7 & 70.2 & 34.0 \\

\midrule
\rowcolor[rgb]{.867, .922, .969}
\multicolumn{5}{c}{\textit{\small\textbf{Skill-based Generators}}} \\
\midrule

DataPrep-Skill (Opus 4.6)
& 7.4 & 34.2 & 70.2 & 35.4 \\

DataPrep-Skill (5.6 Sol)
& 9.2 & 36.1 & 72.8 & 33.6 \\

\textbf{DataFoundry (5.6 Sol)}
& \textbf{13.6} & \textbf{41.3} & 71.1 & \textbf{39.9} \\

\bottomrule
\end{tabular*}
\end{wraptable}

\paragraph{\textsc{DataFoundry}'s prepared datasets generalize across different models}. We replace the downstream base model with Llama-3.1-8B while keeping the training and evaluation settings unchanged. As shown in Table~\ref{tab:synthetic_data_llama}, \textsc{DataFoundry} remains highly competitive across all four domains, achieving the best performance on Math, Finance, and Medical with scores of 13.6, 41.3, and 39.9, respectively. In particular, it outperforms the best data generator by 2.6 points on Math, 4.8 points on Finance, and 3.8 points on Medical. Together with the results above, these findings indicate that the gains of \textsc{DataFoundry} are not tied to a specific model; instead, the prepared data provide transferable training utility across model families.

\subsection{Case Studies}

\begin{figure}
\centering
\vspace{-10pt}
\includegraphics[width=\textwidth]{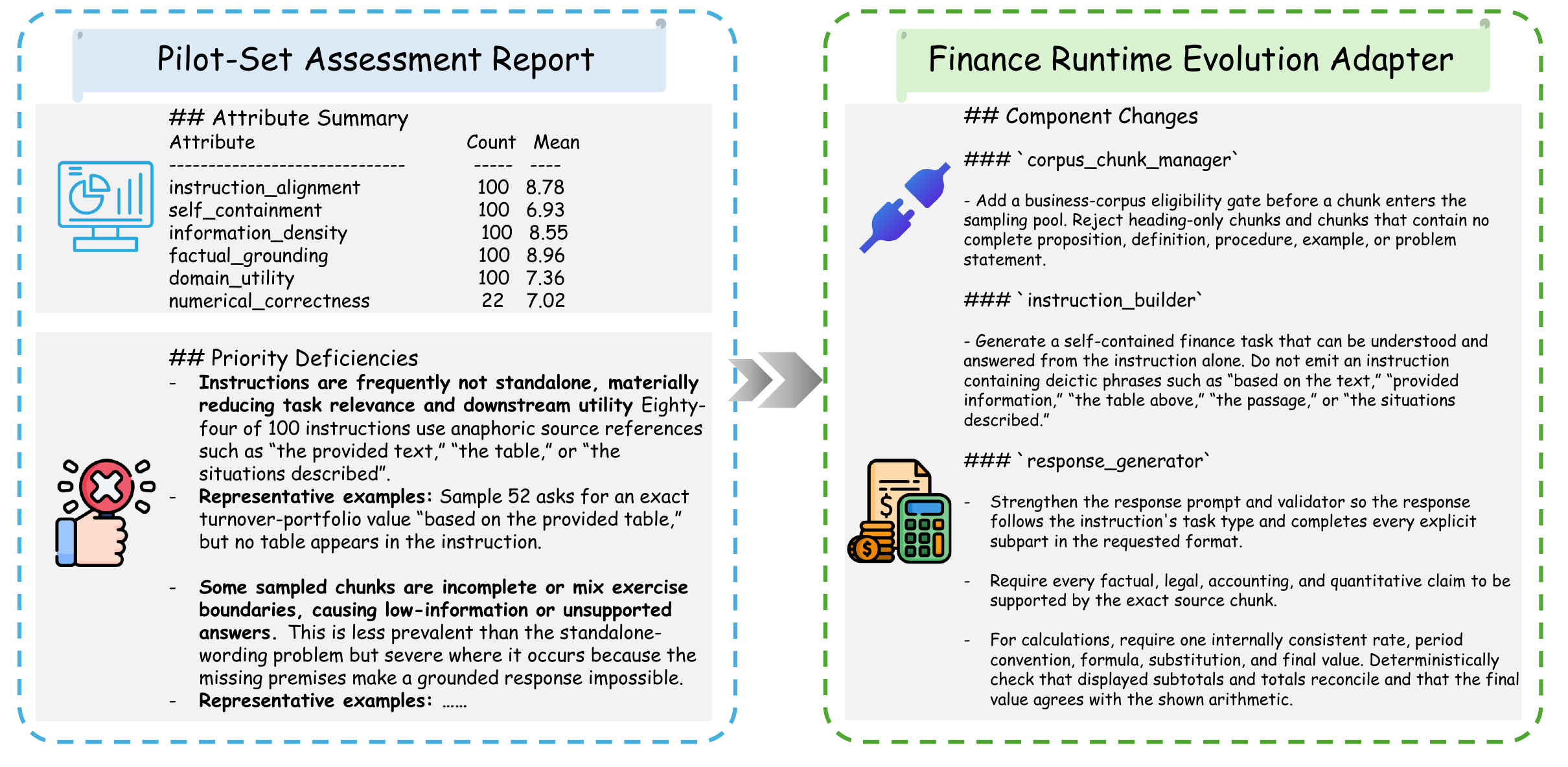}
\caption{A case of implementation of RSI process in \textsc{DataFoundry} in Finance domain.}
\label{fig:RSI_detail}
\end{figure}

\paragraph{\textsc{DataFoundry} translates diagnosed quality deficiencies into targeted component-level improvements.} Figure~\ref{fig:RSI_detail} provides a concrete example of how \textsc{DataFoundry} translates pilot-set deficiencies into targeted component-level revisions. In the Finance domain, the assessment identifies two major sources of quality degradation: many instructions are not self-contained, with 84 of 100 samples relying on missing contextual references such as `the provided text'' or `the table'', making grounded responses difficult. \textsc{DataFoundry} traces the deficiencies back to the responsible runtime components and updates them accordingly. These coordinated revisions illustrate how recursive self-improvement operates at the component level: observed output failures are converted into localized changes to the corresponding stages of the preparation pipeline, progressively specializing the runtime toward domain-specific data-quality requirements. 

\begin{figure}
\centering
\vspace{-10pt}
\includegraphics[width=\textwidth]{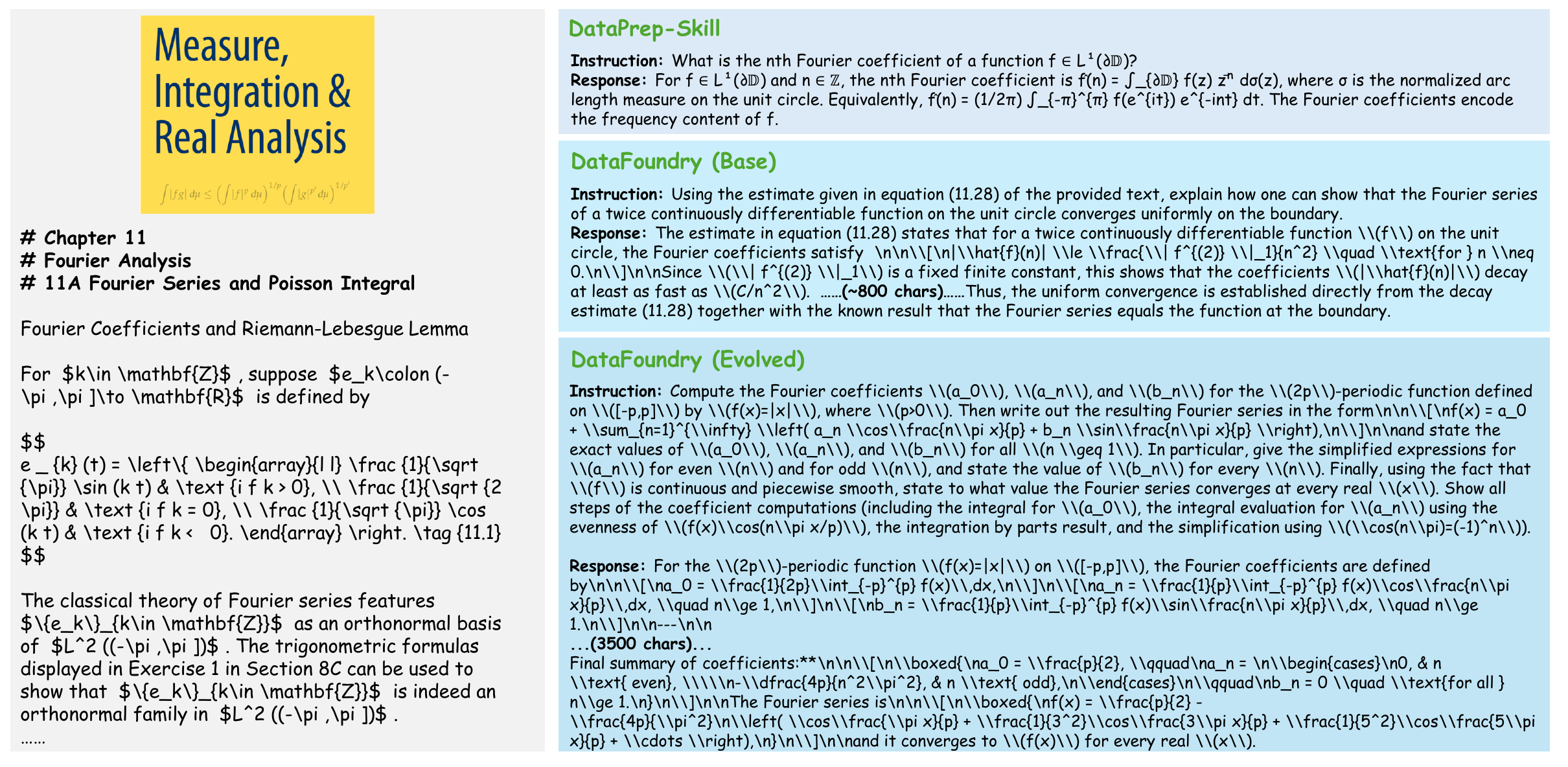}
\caption{Comparison of instances generated by DataPrep-Skill and \textsc{DataFoundry} before and after recursive self-improvement, illustrating progressively improved completeness and self-containment.}
\label{fig:RSI_qa_case}
\vspace{-10pt}
\end{figure}

\paragraph{\textsc{DataFoundry} produces higher-quality training instances, and recursive evolution further improves their quality.}
Figure~\ref{fig:RSI_qa_case} illustrates the progressive improvement in the supervision generated from the same domain corpus. The DataPrep-Skill baseline produces a relatively generic question with limited reasoning requirements, whereas \textsc{DataFoundry} generates substantially richer supervision. Nevertheless, the Base version still exhibits a quality issue identified during assessment: its instruction depends on external context through references such as ``equation (11.28) of the provided text,'' making the example less self-contained. After evolution, \textsc{DataFoundry} transforms the instance into a fully specified Fourier-series problem. The corresponding response provides a complete and directly verifiable solution rather than relying on missing context. This example demonstrates that \textsc{DataFoundry} not only produces more reasoning-intensive training data, but also improves the data by enhancing the completeness and utility of individual examples.
\section{Conclusion}

We present \textsc{DataFoundry}, a framework that improves domain-specific data preparation by recursively evolving the data preparator itself before large-scale synthesis. By diagnosing quality issues on pilot data and translating them into targeted revisions of modular pipeline components, \textsc{DataFoundry} progressively adapts the preparation runtime to domain-specific requirements. Experiments across Math, Finance, Law, and Medical domains show that the resulting datasets consistently provide stronger downstream training utility and generalize across different base models. Our analysis further demonstrates that recursive self-improvement can systematically enhance data quality, while highlighting the need to control excessive evolution.

\newpage
\bibliography{iclr2027_conference}
\bibliographystyle{iclr2027_conference}

\clearpage
\appendix
\section*{Appendix}

\section{Details of Skills-as-Modules Framework}
\label{app:skills}

\subsection{Construction-Runtime-Compiler}

\begin{itemize}[leftmargin=2em]

    \item \textbf{\texttt{corpus\_chunk\_manager}.}
    For each input document, the module partitions the document into
    sections according to its markdown heading hierarchy. It constructs
    content chunks subject to the length constraints
    \texttt{min\_chars}=1200 and \texttt{max\_chars}=3000.
    Sections shorter than the minimum length are merged,
    whereas sections exceeding the maximum length are split.

    \item \textbf{\texttt{instruction\_builder}.}
    Given a content chunk, the module treats the chunk as the knowledge source
    and synthesizes a standalone instruction that is semantically aligned with
    both the source content and its domain. The prompt used in the \textbf{initial specification} is shown
    below:

    \begin{promptbox}[instruction-generation prompt]
<knowledge_context>
{source_chunk}
</knowledge_context>

Create one standalone instruction grounded in the knowledge context. The instruction must be answerable using the context, but must not mention or depend on access to the context. Return only the instruction and do not provide its answer.
    \end{promptbox}

    \item \textbf{\texttt{response\_generator}.}
    The module takes the same content chunk together with the instruction
    produced by \texttt{instruction\_builder} and generates a corresponding
    standalone response. The prompt used in the \textbf{initial specification} is as follows:

    \begin{promptbox}[response-generation prompt]
<knowledge_context>
{source_chunk}
</knowledge_context>

<instruction>
{instruction}
</instruction>

Write one standalone response to the instruction. Ground every factual claim in the knowledge context, but do not mention or depend on access to that context. Return only the response.
    \end{promptbox}

    \item \textbf{\texttt{batch\_generation\_service}.}
    The module concurrently invokes an external language model to generate
    training samples at scale. We use DeepSeek-V4-Flash%
    \footnote{\url{https://api-docs.deepseek.com/}}
    as the external generation model and set the maximum concurrency to 32.

\end{itemize}

\newpage

\subsection{Contextualized-Data-Assessor}

\begin{promptbox}[data-assessment workflow]
1. **Inspect the assessment context.** Resolve the target domain, pilot-set path, and output path.
2. **Load the pilot set.** Read the JSON array.
3. **Select criteria.** Load the library in `multiple_criteria` and explicitly choose one or more criteria that fit the domain and intended downstream use.
4. **Assess by sample.** Judge each candidate independently, one sample at a time using the domain context and the rubric in `multiple_criteria`.
5. **Generate the report.** Aggregate the selected criteria, and identify recurring deficiencies as self-contained dataset-level patterns. The diagnosis stays at the data-quality level for use by a downstream evolution module.
\end{promptbox}

\begin{promptbox}[report-generation format]
# Pilot-Set Assessment Report

## Context
- Domain: ...
- Pilot set: ...
- Selected criteria: ...

## Coverage
- Samples assessed: ...

## Criterion Summary
| Criterion | Count | Mean |
|---|---:|---:|

## Priority Deficiencies
- Self-contained finding describing the affected criterion, breadth, severity, and significance.
  - Representative examples: Sample N - brief excerpt or paraphrase and why it demonstrates the finding; Sample M - brief excerpt or paraphrase and why it demonstrates the finding.
\end{promptbox}

\newpage

\subsection{Evolution-Adapter-Designer}

\begin{promptbox}[adapter-designer workflow]
1. **Review the current setup.** Read the baseline specification, current assessment report, and previous adapter if present. Separate baseline behavior from earlier domain changes.
2. **Select the priority problems.** Compare the recurring deficiencies in the report by prevalence, severity, and likely effect on training-data quality. Choose the problems that warrant changes.
3. **Trace them to modules.** Identify where each problem first appears in the construction process and select the modules responsible for that behavior.
4. **Design the adapters.** State the required module-level changes clearly enough for the runtime compiler to implement.
5. **Carry forward earlier changes.** Merge applicable requirements from the previous adapter into the final effective requirements for each module.
6. **Write the adapter.** Use the structure below.
\end{promptbox}

\begin{promptbox}[report-generation format]
# {Domain} Runtime Evolution Adapter

## Component Changes

### `module_name`
- State the final effective policy, prompt, configuration, or coordination requirements for this module.

## Added Modules
Include this section only when adding modules.

### `module_name`
- Responsibility: ...
- Input: ...
- Output: ...
- Workflow position: ...
- Required behavior: ...
\end{promptbox}

\newpage

\section{Downstream Training Setup}
\label{app:training}

\begin{table}[htbp]
\centering
\caption{Training configuration used in our experiments. We fine-tune all models with LlamaFactory~\citep{zheng2024llamafactoryunifiedefficientfinetuning} using full-parameter supervised fine-tuning (SFT) following~\citep{liang2026dataprepbenchbenchmarkingllmstraining}. Unless otherwise noted, the same configuration is used across all training runs.}
\label{tab:training-hyperparameters}
\small
\renewcommand{\arraystretch}{1.08}
\begin{tabular}{@{}lc@{}}
\toprule
\textbf{Training Parameter} & \textbf{Setting} \\
\midrule
\texttt{fine\_tuning\_type} & Full \\
\texttt{cutoff\_length} & 4{,}096 \\
\texttt{deepspeed\_strategy} & ZeRO-3 \\
\texttt{attention\_backend} & FlashAttention-2 \\
\texttt{precision} & bfloat16 \\
\texttt{batch\_size\_per\_device} & 1 \\
\texttt{gradient\_accumulation} & 4 \\
\texttt{learning\_rate} & $5\times10^{-6}$ \\
\texttt{training\_epochs} & 3 \\
\texttt{lr\_scheduler} & Cosine \\
\texttt{warmup\_ratio} & 0.1 \\
\bottomrule
\end{tabular}
\end{table}

\section{Evaluation Benchmarks}
\label{app:evaluation}

\begin{table}[htbp]
\caption{Downstream evaluation benchmarks used in our experiments, following~\cite{liang2026dataprepbenchbenchmarkingllmstraining}. The released MATH-500 evaluation implementation uses 5,000 examples, on which the reported results were obtained. To avoid ambiguity, we refer to it as MATH-5000 throughout this paper.}
\label{tab:evaluation_benchmarks}
\centering
\small
\renewcommand{\arraystretch}{1.2}
\setlength{\tabcolsep}{3pt}
\begin{tabular}{@{}p{0.17\columnwidth}
p{0.75\columnwidth}
>{\centering\arraybackslash}p{0.08\columnwidth}@{}}
\toprule
\textbf{Benchmark} & \textbf{Description} & \textbf{Count} \\
\midrule
\rowcolor[rgb]{.9,.9,.9}
\multicolumn{3}{c}{\small\textit{\textbf{Mathematics}}} \\
\midrule
GSM8K & Grade-school math word problems requiring multi-step reasoning. & 1,319 \\
AMC 2023 & Multiple-choice problems from the 2023 American Mathematics Competitions. & 40 \\
AIME 2024 & Competition-level mathematics problems from the 2024 American Invitational Mathematics Examination. & 30 \\
MinervaMath & Quantitative reasoning problems drawn from STEM and other technical subjects. & 272 \\
MATH-5000 & A representative datasets of the challenging MATH exam set. & 5000 \\
Gaokao 2024 & Mathematics questions from China's 2024 National College Entrance Examination. & 91 \\
OlympiadBench & Bilingual multimodal olympiad-level mathematics and science problems. & 675 \\
\midrule
\rowcolor[rgb]{.9,.9,.9}
\multicolumn{3}{c}{\small\textit{\textbf{Finance}}} \\
\midrule
CPA-KQA & Expert-authored accounting and finance questions covering CPA-derived concepts. & 210 \\
FinEval-KR & Chinese financial knowledge-and-reasoning questions with annotations. & 101 \\
XFinBench & Multimodal, knowledge-intensive financial problems covering terminology, temporal reasoning, forecasting, planning, and numerical modelling. & 435 \\
\midrule
\rowcolor[rgb]{.9,.9,.9}
\multicolumn{3}{c}{\small\textit{\textbf{Law}}} \\
\midrule
LegalBench & Collaboratively constructed legal-reasoning tasks designed with legal professionals. & 413 \\
LexGLUE & A standardized collection of legal natural-language-understanding datasets. & 7,815 \\
\midrule
\rowcolor[rgb]{.9,.9,.9}
\multicolumn{3}{c}{\small\textit{\textbf{Medical}}} \\
\midrule
MedCaseReasoning & Clinical diagnostic cases paired with clinician-authored reasoning traces. & 897 \\
MedMCQA & Medical entrance-exam multiple-choice questions spanning 21 subjects. & 4,183 \\
MedR-Bench & Structured real-world clinical cases testing examination recommendation, diagnosis, and treatment planning. & 957 \\
\bottomrule
\end{tabular}
\end{table}

\newpage

\section{Baselines Implementation}
\label{app:baselines}

\begin{table}[htbp]
\centering
\caption{Implementation details of the baseline generators from~\citep{liang2026dataprepbenchbenchmarkingllmstraining}.}
\label{tab:baseline-implementations}
{\small
\renewcommand{\arraystretch}{1.15}
\setlength{\tabcolsep}{3pt}
\begin{tabular}{@{}p{0.27\columnwidth}p{0.70\columnwidth}@{}}
\toprule
\textbf{Generator} & \textbf{Operational mechanism} \\
\midrule
\textbf{DataFlow-Based Generator} & DataFlow executes a multi-step workflow, which processes and generates training data through predefined operators. DataFlow-Skill selects or creates DataFlow operators and compose them into an executable pipeline. \\
\midrule
\textbf{LLM-Based Generator} & Prompts an LLM to convert source-book excerpts directly into QA pairs. \\
\midrule
\textbf{Agent-Based Generator} & Runs a ReAct-style agent in a coding harness environment with a standard prompt. The agent iterates until it judges construction complete. \\
\midrule
\textbf{Skill-Based Generator} & Augments the ReAct-style agent in a coding harness environment with an automatically loaded data construction skill. The agent iterates until it judges construction complete. \\
\bottomrule
\end{tabular}
}
\end{table}

\section{Prepared Dataset Size}

\begin{table}[htbp]
\caption{Number of training samples synthesized by each data construction method across domains, all operating on the \emph{same} raw source corpus. The Total column aggregates yield across all domains.}
\label{tab:dataset-sizes}
\centering
\scriptsize
\setlength{\tabcolsep}{6pt}

\begin{tabular*}{\columnwidth}{
    @{\extracolsep{\fill}}
    l|rrrr|r
    @{}
}
\toprule
\small\textbf{Generator} & \small\textbf{Math} & \small\textbf{Finance} & \small\textbf{Law} & \small\textbf{Medicine} & \small\textbf{Total} \\
\midrule
Dolly-15k & - & - & - & - & 15{,}011 \\
\midrule

\rowcolor[rgb]{.867, .922, .969}
\multicolumn{6}{c}{\textit{\small\textbf{DataFlow-based Generators}}} \\
\midrule
DataFlow & 90{,}222 & 111{,}774 & 53{,}891 & 253{,}355 & 509{,}242 \\
DataFlow-Skill & 165{,}345 & 49{,}841 & 21{,}659 & 113{,}305 & 350{,}150 \\
\midrule

\rowcolor[rgb]{.867, .922, .969}
\multicolumn{6}{c}{\textit{\small\textbf{LLM-based Generators}}} \\
\midrule
Claude Opus 4.6 & 2{,}563 & 3{,}750 & 2{,}184 & 7{,}677 & 16{,}174 \\
Gemini 3.0 Pro & 1{,}619 & 1{,}584 & 893 & 2{,}992 & 7{,}088 \\
GPT-5.2 & 9{,}179 & 5{,}832 & 3{,}949 & 14{,}945 & 33{,}905 \\
\midrule

\rowcolor[rgb]{.867, .922, .969}
\multicolumn{6}{c}{\textit{\small\textbf{Agent-based Generators}}} \\
\midrule
Qwen3.5-Plus & 684 & 21{,}693 & 13{,}712 & 25{,}965 & 62{,}054 \\
GLM-4.7 & 684 & 23{,}246 & 13{,}326 & 24{,}734 & 61{,}990 \\
Claude Opus 4.6 & 14{,}892 & 17{,}059 & 19{,}505 & 44{,}284 & 95{,}740 \\
Gemini 3.0 Pro & 5{,}340 & 249 & 389 & 465 & 6{,}443 \\
GPT-5.2 & 14{,}337 & 11{,}999 & 4{,}967 & 30{,}765 & 62{,}068 \\
GPT-5.3-codex & 2{,}500 & 249 & 1{,}584 & 463 & 4{,}796 \\
GPT-5.6 Sol & 13{,}058 & 798 & 1{,}052 & 1{,}216 & 16{,}124 \\
\midrule

\rowcolor[rgb]{.867, .922, .969}
\multicolumn{6}{c}{\textit{\small\textbf{Skill-based Generators}}} \\
\midrule
DataPrep-Skill~(Opus 4.6) & 1{,}439 & 106{,}401 & 24{,}255 & 38{,}351 & 170{,}446 \\
DataPrep-Skill~(GPT-5.6 Sol) & 3{,}501 & 72{,}420 & 34{,}584 & 196{,}747 & 307{,}252 \\
\textbf{DataFoundry~(GPT-5.6 Sol)} & 32{,}692 & 15{,}689 & 17{,}474 & 21{,}711 & 87{,}566 \\
\bottomrule
\end{tabular*}
\end{table}

\newpage

\subsection*{AI use statement}

In this work, we used generative AI tools to generate synthetic datasets, assist in developing theoretical models or conceptual frameworks, design or provide feedback on research methodology or experiments, implement methods, clean or reformat datasets, and assist with translation. We did not use generative AI tools to interpret results, or support qualitative or thematic data analysis. The use of generative AI to formulate mathematical claims, provide critical ingredients for proving mathematical claims, assist in writing proofs, or propose or refine hypotheses was not applicable to this work.

Additionally, we used generative AI tools to draft parts of the paper, summarize and analyze existing literature, identify research topics or gaps, support brainstorming, search for information, improve the readability of the manuscript, identify relevant literature, format references, and suggest possible titles and keywords.

We carefully reviewed all AI-assisted work. Specifically, generative AI was used to synthesize datasets for model training. These synthetic data were reviewed by multiple authors to ensure that they did not contain privacy leakage, harmful content, or other ethical issues. Generative AI was also used to assist in developing the conceptual framework, for example by checking and refining definitions of symbols, notation, and formulas. The resulting framework was independently reviewed and tested by multiple authors for factual correctness and internal consistency.

Generative AI assisted with method implementation, particularly in writing code for automated components of our system. AI-assisted code was reviewed, verified, and tested for correctness by multiple authors. Generative AI was also used for translation and language editing to improve the fluency and naturalness of the manuscript.

For literature-related tasks, generative AI was used to support brainstorming, literature search, and the summarization and analysis of prior work. The authors manually verified relevant literature, source attribution, and the resulting descriptions of prior work, and reviewed the manuscript to avoid unsupported claims or potential plagiarism. Generative AI also assisted with drafting and editing parts of the manuscript, improving readability, formatting references, and suggesting candidate titles and keywords. All AI-generated or AI-assisted text was reviewed and revised by the authors before inclusion in the final manuscript.

We take responsibility for the final content of this work, including all text, claims, code, data, and other artifacts produced with the aid of generative AI.

\end{document}